\documentclass[lettersize,journal]{IEEEtran}
\usepackage{amsmath,amsfonts}
\usepackage{algorithmic}
\usepackage{algorithm}
\usepackage{array}
\usepackage[caption=false,font=normalsize,labelfont=sf,textfont=sf]{subfig}
\usepackage{textcomp}
\usepackage{stfloats}
\usepackage{url}
\usepackage{verbatim}
\usepackage{graphicx}
\usepackage{cite}
\usepackage{caption}

\begin{document}

\title{Bi-VLDoc: Bidirectional Vision-Language Modeling for Visually-Rich Document Understanding}

\author{Chuwei Luo\IEEEauthorrefmark{1}$^{1}$, Guozhi Tang\IEEEauthorrefmark{1}$^{2}$, Qi Zheng$^{1}$, Cong Yao\IEEEauthorrefmark{2}$^{1}$, Lianwen Jin\IEEEauthorrefmark{2}$^{2}$, Chenliang Li$^{1}$, Yang Xue$^{2}$, Luo Si$^{1}$

\IEEEauthorblockA{$^1$Alibaba Group\\
$^2$School of Electronic and Information Engineering, South China University of Technology, China}

\IEEEauthorblockA{chuwei.lcw@alibaba-inc.com, gztang@126.com, yongqi.zq@taobao.com, yaocong.yao@alibaba-inc.com\\
eelwjin@scut.edu.cn, lcl193798@alibaba-inc.com, yxue@scut.edu.cn, luo.si@alibaba-inc.com}

\thanks{\IEEEauthorrefmark{1}These authors contributed equally to this work.}
\thanks{\IEEEauthorrefmark{2}Corresponding authors.}

}

\markboth{Journal of \LaTeX\ Class Files,~Vol.~14, No.~8, August~2021}%
{Shell \MakeLowercase{\textit{et al.}}: A Sample Article Using IEEEtran.cls for IEEE Journals}


\maketitle

\begin{abstract}
Multi-modal document pre-trained models have proven to be very effective in a variety of visually-rich document understanding (VrDU) tasks. Though existing document pre-trained models have achieved excellent performance on standard benchmarks for VrDU, the way they model and exploit the interactions between vision and language on documents has hindered them from better generalization ability and higher accuracy. In this work, we investigate the problem of vision-language joint representation learning for VrDU mainly from the perspective of supervisory signals. Specifically, a pre-training paradigm called Bi-VLDoc is proposed, in which a bidirectional vision-language supervision strategy and a vision-language hybrid-attention mechanism are devised to fully explore and utilize the interactions between these two modalities, to learn stronger cross-modal document representations with richer semantics. Benefiting from the learned informative cross-modal document representations, Bi-VLDoc significantly advances the state-of-the-art performance on three widely-used document understanding benchmarks, including Form Understanding (from 85.14\% to 93.44\%), Receipt Information Extraction (from 96.01\% to 97.84\%), and Document Classification (from 96.08\% to 97.12\%). On Document Visual QA, Bi-VLDoc achieves the state-of-the-art performance compared to previous single model methods.
\end{abstract}


\section{Introduction}
\IEEEPARstart{V}{isually}-rich document understanding (VrDU) aims
at automatically analyzing and extracting important content from visually-rich document images (VRDs) such as forms and receipts. It is an important research area with high academic and application values, since it can make previously labor-intensive and time-consuming workflows for document processing significantly more efficient and convenient. To accurately extract key content from VRDs, textual clues, as well as layout and visual clues all play a vital role. 

\begin{figure}[t]  
\centering
\includegraphics[scale=0.65]{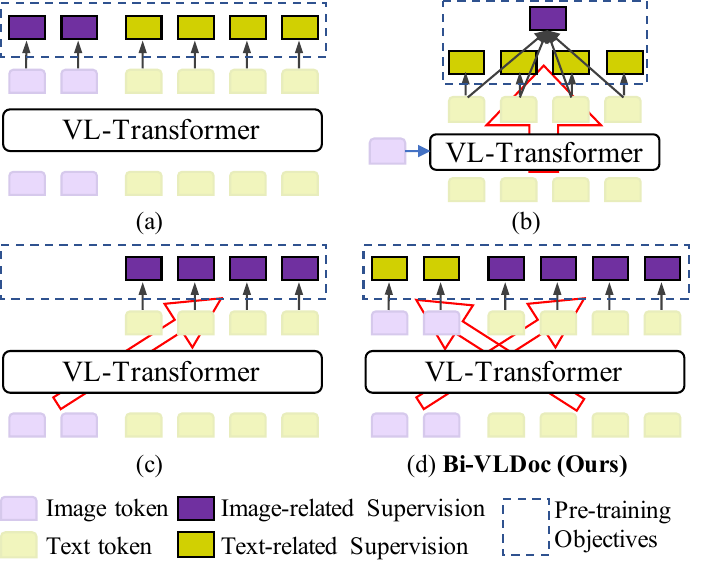}  
\caption{Comparison of multi-modal VrDU pre-trained models. Typical models: (a) SelfDoc\cite{li2021selfdoc} (b) DocFormer\cite{appalaraju2021docformer}  (c) LayoutLM v2\cite{xu2020layoutlmv2}  (d) Bi-VLDoc (Ours). The red arrow means the flow of cross-modality information influenced by the supervisory signals.}
\label{fig:compare_model}
\end{figure}

Recent years have witnessed the success of document pre-trained models in VrDU tasks. By incorporating both text and layout information in transformer-based pre-training architectures, existing methods \cite{xu2020layoutlm, li2021structurallm} work well in various downstream VrDU tasks. To make use of multi-modal information, previous document pre-trained models~\cite{xu2020layoutlmv2,li2021selfdoc,appalaraju2021docformer} have adopted vision-language transformers (VLTs) and explored cross-modal information fusion by introducing vision-language pre-training tasks. 
From the perspective of supervisory signals, vision-language pre-training tasks for VrDU \cite{xu2020layoutlmv2,li2021selfdoc,appalaraju2021docformer} can be roughly divided into three categories: \textbf{\uppercase\expandafter{\romannumeral1}}. corresponding-modality-supervision, \textbf{\uppercase\expandafter{\romannumeral2}}. vision-supervise-language, and \textbf{\uppercase\expandafter{\romannumeral3}}. language-supervise-vision. 
As for type \textbf{\uppercase\expandafter{\romannumeral1}}, although some VrDU pre-trained methods combine multi-modal features into VLTs, the prediction is supervised separately on each modality which means the supervision for the output is basically within its corresponding signal of the same modality, including vision-supervise-vision and language-supervise-language, e.g., MLM (masked language model) in Figure.1 (a). 
As for type \textbf{\uppercase\expandafter{\romannumeral2}}, several methods took a step further to consider the interactions between the modalities in the pre-training task design. More specifically, given tokens and operations on their corresponding image areas, the textual side of the model is imposed with image-related supervision signals, which are generated from operations on the image side, e.g., LTR (learning to reconstruct) \cite{appalaraju2021docformer} in Figure.1 (b) and TIA (text-image alignment) \cite{xu2020layoutlmv2} in Figure.1 (c). The supervisory signals of existing VrDU pre-training tasks mainly belong to types \textbf{\uppercase\expandafter{\romannumeral1}} and \textbf{\uppercase\expandafter{\romannumeral2}}, however, there is limited exploration on type \textbf{\uppercase\expandafter{\romannumeral3}}, where the prediction of the image side is supervised by text-related signals. In summary, due to the design and setup of the pre-training tasks, in most previous cross-modal document pre-trained models, the interactions between vision and language are not fully exploited. Consequently, the leaned joint representations of these models are not sufficiently informative and discriminative.

To tackle this problem, we propose to fully model and make use of the bidirectional interactions between the vision and language modalities in documents for better VrDU (as shown in Figure.1 (d)). In order to achieve this goal, we establish a task-agnostic cross-modal representation learning framework (termed as \textbf{Bi-VLDoc}), in which a vision-language transformer (VLT) is adopted and a bidirectional vision-language supervision strategy as well as a vision-language hybrid-attention mechanism are designed. Specifically, the bidirectional vision-language interaction is enforced by using supervisory signals from the two modalities. Three new pre-training tasks, including a new language-supervise-vision task, a fine-grained layout-related vision-supervise-language task, and a bidirectional vision-language supervised task, are proposed. First, the language-supervise-vision task is achieved by the \textbf{R}egion-\textbf{W}ise \textbf{T}ext \textbf{P}rediction (\textbf{RWTP}) task for better language-to-vision information alignment. Second, since the existing vision-supervise-language task of StructuralLM \cite{li2021structurallm} cannot handle a situation where one text block is distributed across multiple areas, Bi-VLDoc realizes fine-grained position-level layout perception by introducing a pre-training objective called \textbf{T}ext \textbf{I}mage \textbf{P}osition \textbf{A}wareness (\textbf{TIPA}) to make the precise locations of the text blocks in the image to be more accurately aware. Third, Bi-VLDoc presents a new \textbf{B}idirectional \textbf{T}ext-\textbf{I}mage \textbf{A}lignment (\textbf{BTIA}) task that accomplishes vision-supervise-language and language-supervise-vision simultaneously. Moreover, the VLT in Bi-VLDoc is enhanced by vision-language hybrid-attention operations (see Sec. \ref{Sec:HybridAttention} for more details). Extensive experiments on four public benchmarks have demonstrated the effectiveness of the proposed Bi-VLDoc model. 

In summary, the contributions  of this work are summarized as follows:
\begin{itemize}

\item[1)] By using supervisory signals from different modalities to encourage the bidirectional vision-language interaction on documents, this paper introduces bidirectional vision-language supervision for VrDU, which is achieved by three pre-training tasks: RWTP, TIPA, and BTIA.

\item[2)] To learn better document representations with the bidirectional vision-language interaction, this paper introduces a new bidirectional vision-language hybrid-attention mechanism.

\item[3)] Comprehensive experiments were conducted to verify that the proposed Bi-VLDoc achieves top performances on multiple downstream VrDU tasks, outperforming previous state-of-the-art models.

\end{itemize}

\section{Related Work}

\subsection{Visually-rich Document Understanding}
Visually-rich Document Understanding (VrDU) is recent research topic which  aims to automatically handle with the document images. Deep learning approaches have been extensively applied to VrDU tasks like document information extraction and document classification. In visually-rich documents (VRDs), the textual, layout and visual information are crucial for understanding the whole documents. Some methods based on graph neural networks \cite{qian2019graphie,liu2019graph,luo2020merge,li2020end,yu2021pick,tang2021matchvie, Yang, Multimedia} and grid information \cite{katti2018chargrid,denk2019bertgrid} have been proposed to make full use of the textual, layout and image information in VRDs. 
Although these methods have proven to be effective in VrDU, they are heavily task-specific and require extensive annotations. 

\begin{figure*}[htbp]  
\centering  
\includegraphics[scale=0.47]{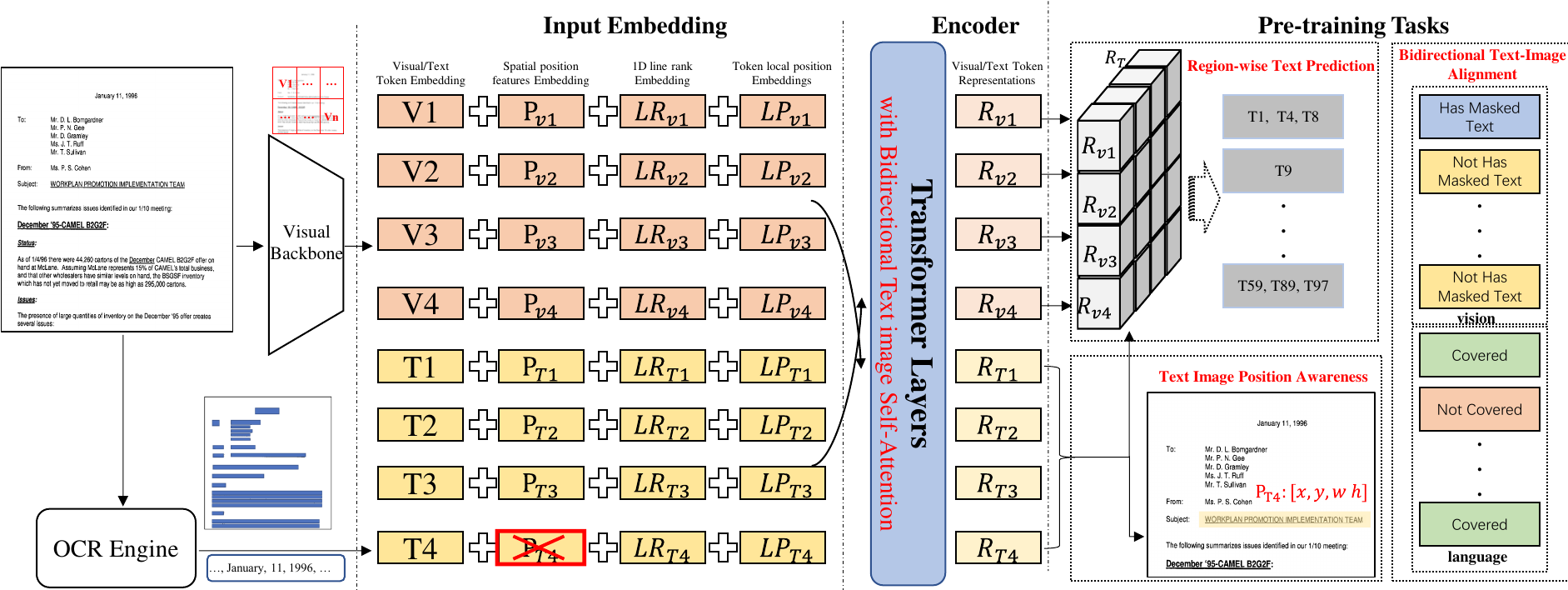}
\caption{Illustration of the model architecture and pre-training strategies for Bi-VLDoc. Bi-VLDoc is a pre-trained multimodal Transformer for VrDu with bidirectional vision-language interaction.  Bi-VLDoc accepts inputs of three modalities, text, image, and layout. The extracted multi-modal embeddings are fed into the bidirectional vision-language hybrid-attention module. Finally, the generated representation can be further utilized for pre-training tasks or downstream VrDU tasks.}
\label{fig:model_overview}
\end{figure*}

\subsection{Vision-Language Pre-training}
Pre-trained models have proven successful in NLP tasks. As the transformer-based pre-trained model, BERT \cite{devlin2019bert} and other models \cite{liu2019roberta,unilmv2} have the ability to learn from the large-scale unlabeled data, and can be adapted to downstream tasks without changing the structure of the model. Inspired by this, some works extend the pre-trained models to vision-language learning \cite{lu2019vilbert,su2019vl,chen2019uniter,li2020unimo,li2020oscar,jia2021scaling,radford2021learning,Word-of-Mouth, Self-adaptiv}. However, these general vision-language models are primarily designed for natural images/videos, thus cannot be directly applied to document related tasks \cite{li2021selfdoc}, since the properties of natural images/videos are quite different from document images. 
Moreover, in document understanding text is \emph{endogenous}, which is a part of the corresponding document, while in general visual understanding tasks such as visual question answering and image captioning, text content is \emph{exogenous} (usually given by human), which is related to the corresponding image/video but not a part of the image/video. Empirical experiments in LayoutLM \cite{xu2020layoutlm} already verified that off-the-shelf pre-trained models cannot work well for VrDU tasks. This is why pre-training models that are specially designed for document understanding are proposed in succession recently.

\subsection{Document Pre-training}
To learn an effective document representation, some works propose pre-trained models for VrDU and achieve state-of-the-art performance in many VrDU downstream tasks.
LayoutLM \cite{xu2020layoutlm} first combines both text and layout information from documents for pre-training. The masked visual-language model (MVLM) is used for pre-training of LayoutLM. 
To avoid focusing on text-only representation, StructuralLM \cite{li2021structurallm} introduces cell-level layout information to make full use of the cell and layout information. 
A vision-supervise-language task named cell position classification (CPC), which aims to predict where the cells are in the documents, is used for pre-training. 
To avoid an excessive contextualization, SelfDoc \cite{li2021selfdoc} uses the semantic components rather than words for multi-modal pre-training. The supervisory signals of SelfDoc are from the corresponding modality.
In addition to multi-modal attention, DocFormer \cite{appalaraju2021docformer} introduces pre-training tasks which are designed by vision-supervise-language and corresponding-modality supervision. 
To extend the vanilla LayoutLM, LayoutLM v2 \cite{xu2020layoutlmv2} integrates the image information and a spatial-aware self-attention mechanism to learn the relative positional relationship. 
The pre-training tasks of LayoutLM v2 contains MVLM, TIA and text-image matching (TIM). The supervisory signals of them are from vision and language but only on the text side.
Although most of the existing VrDU methods employ multi-modalities during pre-training, the supervisory signals of existing VrDU pre-training tasks mainly come from corresponding-modality-supervision and vision-supervise-language.
Pre-training tasks that use the language-supervise-vision have not been fully studied. Thus, these VrDU models are not bidirectionally cross-modal which means that correlation of vision and language cannot be well learned.

\section{Bi-VLDoc}
This section presents Bi-VLDoc, a bidirectional visual-linguistic document pre-trained model for visually-rich document understanding. The overall framework of Bi-VLDoc is shown in Figure \ref{fig:model_overview}. Bi-VLDoc achieves the bidirectional vision-language interaction by introducing not only a bidirectional hybrid-attention transformer mechanism, but also bidirectional vision-language supervision. Specifically, to emphasize features in language and vision for multi-modal fusion, the BVLHA (bidirectional vision-language hybrid-attention) module is proposed to contribute a cross-modal feature extraction encoder. To more accurately utilize the layout position information, the TIPA (text image position awareness) task is presented to simulate the bounding box regression task and make full use of the anchor box representation. To further explore the design of language-supervise-vision tasks, the RWTP (region-wise text prediction) task is introduced. Furthermore, to emphasize the bidirectional perception of both vision and language, the BTIA (bidirectional text-image alignment) task is utilized to execute the language-supervise-vision and vision-supervise-language tasks simultaneously.

\subsection{Model Architecture} The overview of the proposed Bi-VLDoc is shown in Figure \ref{fig:model_overview}. Similar to LayoutLM v2 \cite{xu2020layoutlmv2}, Bi-VLDoc accepts inputs of three modalities, text, image, and layout. The extracted multi-modal embeddings are fed into the bidirectional vision-language hybrid-attention module. Finally, the generated representation can be further utilized for pre-training tasks or downstream VrDU tasks.

\subsection{Input Embedding}
\label{sec:inputembedding}

The settings related to text token embedding $\hat{T}=(t_1, ..., t_n)$ are the same as \cite{li2021structurallm}, where $n$ represents the number of text tokens, set to 512.. We follow LayoutLM v2 \cite{xu2020layoutlmv2} to get visual token embeddings $\hat{V}=(v_1, ..., v_m)$, where $m$ represents the number of visual tokens, set to 49. For spatial position features $\hat{P}$, we introduce the token local position embedding $LP_{emb}$  to represent the relative position of the token in each text segments. In detailed, given $i$-th text segment in a document, the internal token local position embedding $LP_{emb}$ for the $i$-th text segment can be represented as $\{1,2,...,2, 3\}$, here, 1 indicates the begin token of the text segment, 2 indicates the middle token of the text segment, and 3 indicates the end token of the text segment. Other spatial position settings are the same as \cite{li2021structurallm}. The final text embedding $H_t=\{h_{t_1}, ..., h_{t_n}\}$ is the sum of text token embeddings and all positional embeddings. The final visual token embedding $H_v=\{h_{v_1}, ..., h_{v_m}\}$ is the sum of visual token embeddings and all positional embeddings. 


\begin{align}
H_{vt}^l &= \{h^l_{v_1}, ..., h^l_{v_m}, h^l_{t_1}, ..., h^l_{t_n}\} \\
&= Transformer_l(H_{vt}).
\end{align}
We pack the final text token embeddings and the final visual token embeddings together into $H_{vt}=\{h_{v_1}, ..., h_{v_m}, h_{t_1}, ..., h_{t_n}\}$. An $L$-layer vision-language Transformer encoder is used to get the vision-language output.
\subsection{Bidirectional Vision-Language Hybrid-Attention} \label{Sec:HybridAttention}

Most previous multi-modal works \cite{jain2019multimodal} build their encoder with a stack of multi-head self-attention layers followed by a feed-forward network which come from \cite{devlin2019bert}, and thus the multi-modal features have been fused with a proportional weight before being fed to the encoder. However, when the semantics of the text segments are similar, the model needs a stronger emphasis on visual clues, while the opposite applies. Therefore, the general idea is to apply sample dependent attention weights to the different modalities. Based on SelfDoc \cite{li2021selfdoc}, we additionally introduce the fusion mechanism which is named bidirectional vision-language hybrid-attention mechanism.

The output of the $L$-layer vision-language Transformer $H_{VT}^l$ is then split into the new visual representations $H_v^l=\{h^l_{v_1}, ..., h^l_{v_m}, 0_1, ..., 0_n\}$ and the new text representations $H_t^l=\{0_1,...,0_m,h^l_{t_1}, ..., h^l_{t_n}\}$. Eqs. (3-5) are used to combine vision-for-language information with language self-attended information. The intuitive idea is that if the semantics of the text segments are similar, such as handwritten documents, the vision should be accentuated to help better distinguish the multi-modal features.

\begin{align}
H^l_{t\_t} ={\rm softmax}(\frac{q(H^l_{t})k(H^l_{t})^{T})}{\sqrt{d_{k}}})v(H^l_{t})\cdot M_t,\\
H^l_{v\_t} = {\rm softmax}(\frac{q(H^l_{t})k(H^l_{v})^{T})}{\sqrt{d_{k}}})v(H^l_{v})\cdot M_v, \\
H^l_{hybrid-t}=\mathrm{LN}(H^l_{t\_t} + H^l_{v\_t}),
\end{align}

\noindent{where} $M_t=\{0_1,...,0_m,1_1,...,1_n\}$ denotes the text attention mask. $M_v=\{1_1,...,1_m,0_1,...,0_n\}$ denotes the vision attention mask. $H_{t\_t}=\{0_1,...,0_m,h^l_{t\_t_1}, ..., h^l_{t\_t_n}\}$ is language self-attended information. $H_{v\_t}=\{0_1,...,0_m,h^l_{v\_t_1}, ..., h^l_{v\_t_n}\}$ is the vision-for-language information. $H_{hybrid-t}=\{0_1,...,0_m,h^l_{hybrid-t_1}, ..., h^l_{hybrid-t_n}\}$ is the hybrid attention of the text. $LN$ denotes layer normalization. 

On the other side, if the visual features of the text segments are similar, such as a printed digital receipt, the semantic information should be accentuated to help better distinguish the multi-modal features. Therefore, by combining language-to-vision information with vision self-attended information, we can obtain $H_{hybrid-v}$ which is the hybrid attention of the vision side, in Eqs. (6-8),

\begin{align}
H_{v\_v}={\rm softmax}(\frac{q(H^l_{V})k(H^l_{v})^{T})}{\sqrt{d_{k}}})v(H^l_{v})\cdot M_v,\\
H_{t\_v} = {\rm softmax}(\frac{q(H^l_{v})k(H^l_{t})^{T})}{\sqrt{d_{k}}})v(H^l_{t})\cdot M_t,\\
H_{hybrid-v}=\mathrm{LN}(H_{v\_v} + H_{t\_v}),
\end{align}

\noindent{where} $H_{v\_v}=\{h^l_{v\_v_1}, ..., h^l_{v\_v_m},0_1,...,0_n\}$ is vision self-attended information. $H_{t\_v}=\{h^l_{v\_t_1}, ..., h^l_{v\_t_n},0_1,...,0_n\}$ is the language-to-vision information. $H_{hybrid-v}=\{h^l_{hybrid-v_1}, ..., h^l_{hybrid-v_m},0_1,...,0_n\}$ is the hybrid attention of the text. $LN$ denotes layer normalization. 

Finally, the hybrid features that contain both language-to-vision and vision-to-language information are combined to get the bidirectional vision-language output $H_{hybrid}$ of the module.

\begin{equation}
\begin{aligned}
H_{hybrid}&=\mathrm{LN}(H_{hybrid-t} + H_{hybrid-v})\\
&= \{h_{hybrid-v_1}, ..., h_{hybrid-v_m}, \\
& \qquad h_{hybrid-t_1}, ..., h_{hybrid-t_n}\}.
\end{aligned}
\end{equation}

\subsection{Bidirectional Vision-Language Supervision} 
The proposed bidirectional vision-language supervision is achieved by three new pre-training tasks, including a vision-supervise-language task (TIPA), a language-supervise-vision task (RWTP), and a bidirectional text-image supervised task (BTIA).

\begin{figure*}[htbp]  
\centering  
\includegraphics[scale=0.7]{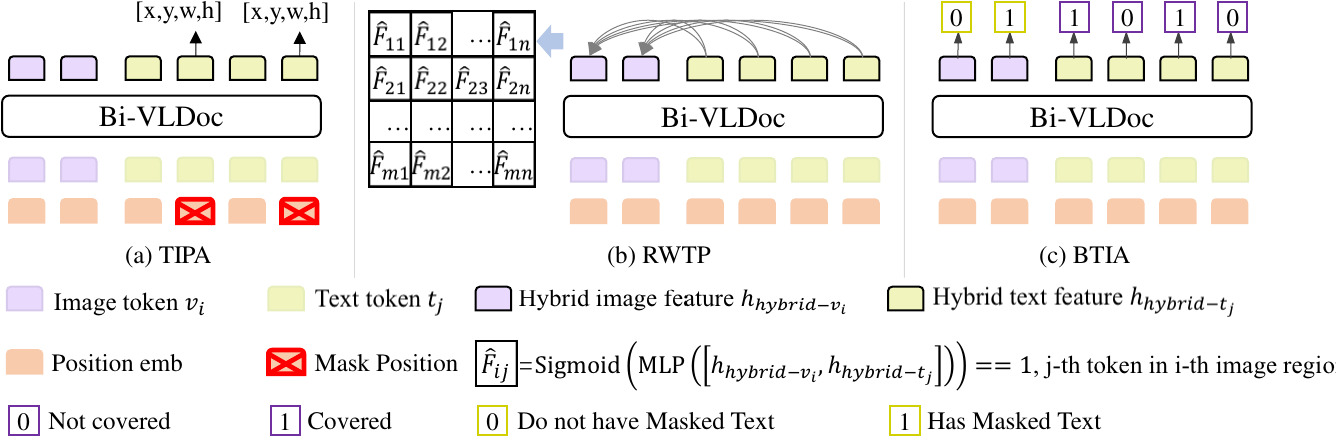}
\caption{Bidirectional vision-language supervision achieved by three pre-training tasks in Bi-VLDoc.}
\label{fig:pre_training_tasks}
\end{figure*}

\noindent{\textbf{Text Image Position Awareness:}} In order to make the document model better understand layout information during pre-training, designing layout-related vision-supervise-language tasks for the language side of the vision-language document pre-trained model is proved to be effective. The cell position classification (CPC) task proposed by StructuralLM \cite{li2021structurallm} works well for pre-training models to understand layout information. By dividing the document image into several regions of the same size, the CPC task aims to predict the text belongs to which region of the document image.

However, like the CPC task, existing layout-related vision-supervise-language tasks are usually with the coarse-grained area classification which cannot handle a situation in which one text block is distributed across multiple regions. It is difficult for the text tokens to understand its layout information precisely.
To address this issue, motivated by text detection methods in computer vision, this work proposes a new vision-supervise-language task, named text image position awareness task (TIPA), to model the relative spatial position of cells in a document. Given a set of scanned documents, TIPA aims to more precisely predict the position of cells in the documents.

Different from the text detection methods in computer vision that predict bounding boxes from the image, in the TIPA (in Figure \ref{fig:pre_training_tasks} (a)), the Bi-VLDoc model predict the bounding box of text token from the language side. This is a typical vision-supervise-language task. The textual side of the Bi-VLDoc model is imposed with image-related supervision signals.

In the TIPA, some percentage of the input spatial positions of the input tokens from document text lines are masked at random (set to [0,0,0,0]), and then predicting only those masked  spatial positions. By split the encoder outputs $H_{hybrid}$ into the final hybrid textual representation $H_{hybrid-t}=\{0_1,...,0_n,h_{hybrid-t_1}, ..., h_{hybrid-t_n}\}$ and the final visual representation $H_{hybrid-v}=\{h_{hybrid-v_1}, ..., h_{hybrid-v_m},0_1,...,0_n\}$, a fully-connected layer is then built above the the final hybrid textual representation $H_{hybrid-t}$ to calculate a vector that represents the four parameterized coordinates. The Distance-IoU Loss \cite{distanceiOU} is utilized to minimize the normalized distance between central points of predicted boxes and target boxes 
to get the $\mathcal{L}_{TIPA}$:

\begin{align}
\mathcal{L}_{TIPA}=1 -{\bf IoU} + \frac{\rho ^2({\bf MLP}(H_{hybrid-t}), b^{gt})}{c^2}),
\end{align}

\noindent{where} $b^{gt}$ denotes the central points of bounding boxes, $\rho$(·) is the Euclidean distance, and c is the diagonal length of the smallest enclosing box covering the two boxes.

\noindent{\textbf{Region-wise Text Prediction:}} Intuitively, it is reasonable to believe that a document understanding model that makes full use of the cross-modalities supervision is more powerful than either a vision-supervise-language VrDU model or the model that supervised by the corresponding modality. However, as aforementioned, the designs of language-supervise-vision are not fully explored in previous VrDU research. To address this limitation, motivated by text recognition methods in computer vision, region-wise text prediction (RWTP) task is introduced.

To further explore language-to-vision understanding, as shown in Figure \ref{fig:pre_training_tasks} (b), the RWTP predicts the relationship of a document's image to its text. Different from the text recognition methods in computer vision that predict text tokens from the input text line image, in the RWTP, the Bi-VLDoc model predict the all text tokens contained in each visual token (region).
Each visual token is supervised by the text information that is contained in the corresponding visual region. This is a typical language-supervise-vision task. The visual side of the Bi-VLDoc model is imposed with language-related supervision signals.

The RWTP is modeled as a binary classification problem to predict which text is in a piece of the image. Specifically, as mentioned before, the image is divided into $m$ areas ($m$ visual tokens) of the same size according to the shape of the visual feature before input to the $L$-layer transformers in Bi-VLDoc. During the pre-training, given the final output textual representations $H_{hybrid-t}$  and visual representations $H_{hybrid-v}$, the representations are aggregated according to each other’s dimension. 

\begin{align}
\!F \!=\! 
\begin{bmatrix}
\!(h_{hybrid-v_1}, \!h_{hybrid-t_1})\!, \!&\!...\!&\!(h_{hybrid-v_1}, h_{hybrid-t_n})\! \\ 
\!... \!&  &\! ...\\ 
\!(h_{hybrid-v_m}, h_{hybrid-t_1})\!,  \!&\!...\!&\!(h_{hybrid-v_m}, \!h_{hybrid-t_n})\! .
\end{bmatrix}
\end{align}
 
Then, an MLP layer with the Sigmoid activation is applied to calculate the combinational representations for representing a text token in/not in a visual token. 

\begin{equation}
\hat{F} = \bf{Sigmoid}(\bf{MLP}(F)),
\end{equation}
\noindent{where} $\hat{F}$ represents the predicted probability distribution, which predicts a label for each area of visual features depending on whether a text is in a piece of the image with a cross-entropy loss $\mathcal{L}_{RWTP}$.
\begin{align}
\mathcal{L}_{RWTP}=\sum_{i=0}^m -(y_{ij} \times log(\hat{F}) + (1-y_{ij}) \times log(1-\hat{F})),
\end{align}
 \noindent{where}  $y_{ij}$  is the label.  For example, if $y[i,j]==1$, this means the $j$-th text token is located in the $i$-th visual regions (tokens). 

As for the label generation, given the divided visual features, the label is generated according to the threshold of the IOU (Intersection over Union) map between each image region and the text box. If the IOU is higher than a threshold (0.5), the label is set to 1; otherwise, 0.

\noindent{\textbf{Bidirectional Text-Image Alignment:}} Beyond the designs that are only on language-supervise-vision or vision-supervise-language task, as shown in Figure \ref{fig:pre_training_tasks} (c), a new bidirectional text-image alignment (BTIA) task is introduced to combine TIA (text-image-alignment) and ITA (image-text-alignment) simultaneously for bidirectional vision-language supervision. 

In the classic TIA \cite{xu2020layoutlmv2} task, given an image with some text regions randomly covered, a classification layer is used for predicting a label for each text token depending on whether it is covered in the document image. This is a vision-supervise-language process that encourages the visual and layout information into text information. However, it ignores the language-supervise-vision process in which visual and layout features can be aware of textural features.

Intuitively, it is reasonable to believe that a bidirectional operation is better than an only vision-supervise-language supervision. Therefore, motivated by the TIA task, ITA is proposed to extend the design on the visual side. In the MVLM task, some text tokens are randomly masked. During the pre-training, a a classification layer is built above the final visual representations $H_{hybrid-v}$. It predicts a label for each area of visual features depending on whether it has the masked text tokens in MVLM, i.e., \textbf{[Has Masked Text]} or \textbf{[Do Not Have Masked Text]}.

By combining both TIA task and ITA task, a new bidirectional text-image alignment (BTIA) task is introduced. Since the final visual features and text features have the same dimensions, to perform deep bidirectional modeling, the visual and textual side of Bi-VLDoc share a same MLP classifier in the BTIA as follows:

\begin{align}
\hat{F}_{BTIA} = {\bf Sigmoid}({\bf MLP}(H_{hybrid})),
\end{align}

{\noindent}where $\hat{F}_{BTIA}$ represents the classification results. $\hat{F}_{BTIA}[0:m]$ denotes the results of ITA task. $\hat{F}_{BTIA}[m+1:m+n]$ denotes the results of TIA task. As for the ITA label generation, following RWTP, given the divided visual features, the label is generated according to the threshold of the IOU (Intersection over Union) map between each image region and the text box. If the IOU is higher than a threshold (0.5), the text token is in this image region; otherwise, not in this image region. The ITA label generation is following LayoutLM v2\cite{xu2020layoutlmv2}.

Therefore, the loss function $\mathcal{L}_{BTIA}$ of BTIA task is defined as follows:

\begin{align}
\mathcal{L}_{BTIA}=\sum_{i=0}^m \log (\hat{F}_{BTIA}, Y^{i}),
\end{align}

{\noindent}where $Y^{i}$  is the matching label.
\section{Experiments}

\subsection{Datasets}



The proposed Bi-VLDoc model is pre-trained on the wildly-used document understanding pre-training dataset. We evaluate Bi-VLDoc on four datasets from four document understanding tasks: form understanding, receipt IE, document VQA, and document  classification. Details are as follows.

\textbf{Pre-training Dataset} Following LayoutLM v2 \cite{xu2020layoutlmv2}  and StrucutralLM \cite{li2021structurallm}, we pre-train Bi-VLDoc on the IIT-CDIP Test Collection 1.0\cite{lewis2006building}. It is a large-scale
scanned document image dataset that are wildly-used for document understanding pre-training. It contains over 11 million scanned document images. Following StrucutralLM \cite{li2021structurallm}, we extract
text and corresponding bounding boxes from document page images by using the open source OCR engine Tesseract\footnote{\url{https://github.com/tesseract-ocr/tesseract}}.

\textbf{FUNSD} FUNSD\cite{jaume2019funsd} is a form document understanding dataset. The dataset consist of 199 real, fully annotated, scanned forms with 9,707 semantic entities and 31,485 words. The training set includes 149 forms, and the test set includes 50 forms. Like other document understanding models, we use the official OCR annotation and fine-tune Bi-VLDoc on semantic entity labeling of FUNSD. Specifically, one of the four predefined categories(question, answer, header or other) is assigned to each word in the dataset. Following the previous works\cite{li2021structurallm}, the word-level F1 score is used for evaluation.

\textbf{CORD} CORD\cite{park2019cord} is a dataset for receipt key information extraction in real world scanned documents. It contains 800 receipts for training, 100 for validation, and 100 for testing. Following LayoutLM v2 \cite{xu2020layoutlmv2} , we ues the official OCR annotations and the ROI as input instead of the raw photo. The task is to label each word to the right field from the defined 30 fields under 4 categories. The entity-level F1 score is used for evaluation.

\textbf{RVL-CDIP} RVL-CDIP\cite{harley2015evaluation} is a dataset for document classification. It consists of 400,000 grayscale images. The dataset is split in an 80:10:10 ratio for training, validation, and testing. There are 16 classes, with 25,000 images per class in the dataset. It is a multi-class single-label classification task, including letter, form, invoice, etc. The overall classification accuracy is used as the evaluation metric. Following StructuralLM, we extracte text and layout information by Tesseract OCR.

\textbf{DocVQA} DocVQA\cite{mathew2021docvqa} is a dataset for visual question answering in document understanding field. The dataset consists of 50,000 questions defined on 12,000+ document images. The dataset is split in an 80:10:10 ratio to train, validation and test splits. Text and layout information is extracted by a public OCR engine. ANLS (aka average normalized Levenshtein similarity) that is based on edit distance is used for evaluation. We get results on the test set by the official evaluation site.

\begin{table}
\centering
\caption{Ablation study on the DocVQA validation set. "BVLHA" means the bidirectional vision-language cross-attention. "BTIA", "RWTP" and "TIPA" are three proposed pre-training objectives.}
\label{tab:ablation}
\begin{tabular}{lccccc}
\hline\hline
\textbf{\#} & \textbf{BVLHA} & \textbf{BTIA} & \textbf{RWTP} & \textbf{TIPA} & \textbf{ANLS} \\
\hline
1  & & & & & 0.7886\\\hline
2  & $\checkmark$ & & & & 0.8050\\\hline
3a & $\checkmark$ & $\checkmark$ & & & 0.8131\\
3b & $\checkmark$ &  & $\checkmark$ & & 0.8126\\
3c & $\checkmark$ &  &  & $\checkmark$ & 0.8175\\
3d & $\checkmark$ & $\checkmark$ & $\checkmark$ &  & 0.8183\\
3e & $\checkmark$ & $\checkmark$ &  & $\checkmark$ & 0.8246\\
3f & $\checkmark$ &  & $\checkmark$ & $\checkmark$ & 0.8265\\\hline
4  & $\checkmark$ & $\checkmark$ & $\checkmark$ & $\checkmark$ & 0.8349\\
\hline\hline
\end{tabular}

\end{table}
\subsection{Pre-training Settings}

\begin{table*}[htb!]

\caption{Model accuracy (entity-level Precision, Recall, F1) on the test set of FUNSD. }
\label{tab:funsd}
\centering
\begin{tabular}{lcccc}
\hline\hline
\textbf{Model} & \textbf{Precision} & \textbf{Recall} & \textbf{F1} & \textbf{Parameters}  \\
\hline
BERT$_{LARGE}$ \cite{devlin2019bert}                & 0.6113 & 0.7085 & 0.6563 & 340M \\
UniLMv2$_{LARGE}$ \cite{unilmv2}             & 0.6780 & 0.7391 & 0.7072 & 355M\\
BROS \cite{hong2020bros}                            & 0.8056 & 0.8188 & 0.8121 & 139M \\\hline
SelfDoc \cite{li2021selfdoc}                        & -      & -      & 0.8336 & 137M \\
DocFormer$_{LARGE}$ \cite{appalaraju2021docformer}  & 0.8229 & 0.8694 & 0.8455   & 536M       \\
LayoutLM$_{LARGE}$ \cite{xu2020layoutlm}            & 0.7596 & 0.8219 & 0.7895  & 343M\\
LayoutLM v2$_{LARGE}$ \cite{xu2020layoutlmv2}        & 0.8324 & 0.8519 & 0.8420   &  426M     \\
StructuralLM$_{LARGE}$ \cite{li2021structurallm}    & 0.8352 & 0.8681 & 0.8514  & 355M \\\hline
Bi-VLDoc                & 0.9060 & 0.9646 & \textbf{0.9344} & 409M \\\hline
\end{tabular}

\end{table*}

\begin{table}[]
\caption{Classification accuracy on the RVL-CDIP test set.}
\label{tab:rvlcdip}
\centering

\begin{tabular}{lc}
\hline\hline
\textbf{Model} & \textbf{Accuracy} \\
\hline
BERT$_{LARGE}$          & 89.92\% \\
UniLMv2$_{LARGE}$       & 90.20\% \\\hline
VGG-16$^{a}$\cite{afzal2017cutting}                 & 90.97\% \\\hline
SelfDoc$_{LARGE}$      & 92.29\%          \\
LayoutLM$_{LARGE}$      & 94.43\%          \\
DocFormer$_{LARGE}$      & 95.50\%          \\
LayoutLM v2$_{LARGE}$    & 95.64\%          \\
StructuralLM$_{LARGE}$   & 96.08\% \\\hline
Bi-VLDoc                 & \textbf{97.12\%} \\\hline\hline
\end{tabular}

\end{table}

\begin{table}[htb!]
\caption{Model accuracy (Precision, Recall, F1) on the test set of CORD.}
\label{tab:cord}
\centering
\begin{tabular}{lccc}
\hline\hline
\textbf{Model} & \textbf{Precision} & \textbf{Recall} & \textbf{F1} \\
\hline
BERT$_{LARGE}$          & 0.8886 & 0.9168 & 0.9025 \\
UniLMv2$_{LARGE}$       & 0.9123 & 0.9289 & 0.9205 \\
BROS                    & 0.9558 & 0.9514 & 0.9536 \\
LayoutLM$_{LARGE}$      & 0.9432 & 0.9554 & 0.9493          \\
LayoutLM v2$_{LARGE}$    & 0.9565 & 0.9637 & 0.9601          \\
DocFormer$_{LARGE}$     & 0.9725 & 0.9674 & 0.9699 \\\hline
Bi-VLDoc                 & 0.9831 & 0.9739 & \textbf{0.9784} \\\hline\hline
\end{tabular}

\end{table}

\begin{table}

\centering
\caption{Average Normalized Levenshtein Similarity (ANLS) score on the DocVQA test set (until 2022-06-27). }
\label{tab:docvqa}
\begin{tabular}{lc}
\hline\hline
\textbf{Model} & \textbf{ANLS} \\
\hline
BERT$_{LARGE}$          & 0.6745 \\
RoBERTa$_{LARGE}$       & 0.6952 \\
UniLMv2$_{LARGE}$       & 0.7709 \\\hline
LayoutLM$_{LARGE}$      & 0.7259          \\
LayoutLM v2$_{LARGE}$ (single model)   & 0.8672          \\
StructuralLM$_{LARGE}$ (single model)  & 0.8394 \\
TILT(single model)\cite{powalski2021going}               & 0.8705\\\hline
Bi-VLDoc                       & \textbf{0.8798} \\\hline
Top-1 on DocVQA Leaderboard (multi-models)   & \textbf{0.8866} \\\hline\hline
\end{tabular}

\end{table}
\textbf{Pre-training Bi-VLDoc} The proposed Bi-VLDoc model is based on the typical large Transformer-based-language-models. It consists of a 24-layer 16-head Transformer encoder with 1024 embedding/hidden size and 4096 feed-forward filter size. There is only one cross attention layer connected the Transformer encoder. In the cross attention layer, the self-attention of each modality layer is 16-head and 1024-hidden-size. 

Following previous works\cite{Baralis, xu2020layoutlmv2}, we initialize the weight of Bi-VLDoc model with the existing pre-trained models. For 24-layer transformer encoder layers and the text embedding layer, we use the RoBERTa-Large\cite{liu2019roberta} to initialize. For the visual part in the Bi-VLDoc, we use the backbone of a Mask R-CNN\cite{he2017mask} model trained on PubLayNet\cite{zhong2019publaynet} following LayoutLM v2\cite{xu2020layoutlmv2}. The cross attention layer and the rest of the parameters in the model are randomly initialized. The Bi-VLDoc is trained by using Adam optimizer with the learning rate of 2e-5 and a linear
decay learning rate schedule. We use batch size 384 to train Bi-VLDoc for 10 epochs on the IIT-CDIP dataset.

For pre-training on IIT-CDIP dataset, we set the maximum sequence length to 512 and keep the head and the tail of the text sequence if the text is too long. All text tokens are set to the segment 0 and visual feature tokens are set to the segment 1. The output of the visual backbone is transformed by a pooling layer to 49 ($7\times7$) image tokens. Following \cite{devlin2019bert}, we mask 15\% text tokens in which 80\% are replaced by the [MASK] token , 10\% are replaced by a random token, and 10\% keeps unchanged. In BTIA, 15\% text blocks are covered and the visual parts performs image-text-alignment according to the text blocks that contains [MASK] token from MVLM. In TIPA, the 2D-positions of 15\% text blocks are masked by (0,0,0,0).

\subsection{Ablation Study}
To better understand the contributions of different components of Bi-VLDoc, following LayoutLM v2 \cite{xu2020layoutlmv2}, we perform ablation studies on the DocVQA validation set. The ground-truth OCR text is used to avoid the wrong OCR text from interfering with our ablation analysis. Considering that the complete pre-training takes too long, we pre-train Bi-VLDoc using the subset of IIT-CDIP that contains one million documents for 3 epochs.

Table \ref{tab:ablation} shows model performance of Bi-VLDoc on the DocVQA validation set. By comparing \#1 and \#2, we investigate the effectiveness of Bidirectional Vision-Language Hybrid-Attention (BVLHA) in our model. The results demonstrate that introducing the BVLHA can obtain better performance by adapting the diverse variety of document images. we conduct an ablation study in \#3a, \#3b and \#3c, to explore the effect of each designed pre-training task. From the results in \#3a, \#3b and \#3c, we observe all of the introduced tasks improving the model performance substantially, and the proposed TIPA task even more than the others. From the comparison results of \#3d, \#3e and \#3f, we combine the proposed pre-training tasks in pairs to evaluate the effect of task combination. We notice that using both tasks together is more effective than using either one alone.  Other than that, Bi-VLDoc demonstrates the best performance when all the proposed modules are used for pre-training.


\begin{figure}[htb!]
\centering
\includegraphics[scale=0.9]{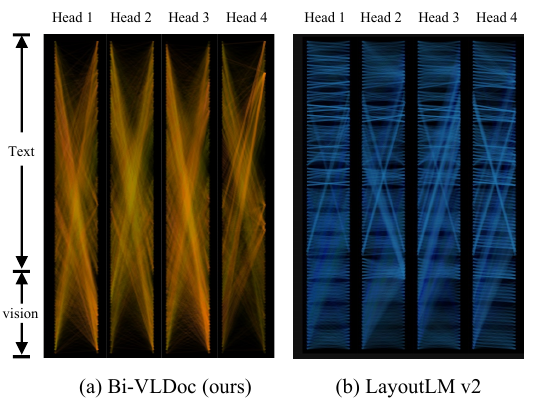}
\caption{Comparison of the first four attention heads between our proposed Bi-VLDoc and LayoutLM v2 \cite{xu2020layoutlmv2}}
\label{fig:vis}
\end{figure}

\begin{figure}[htb!]
\centering
\includegraphics[scale=0.45]{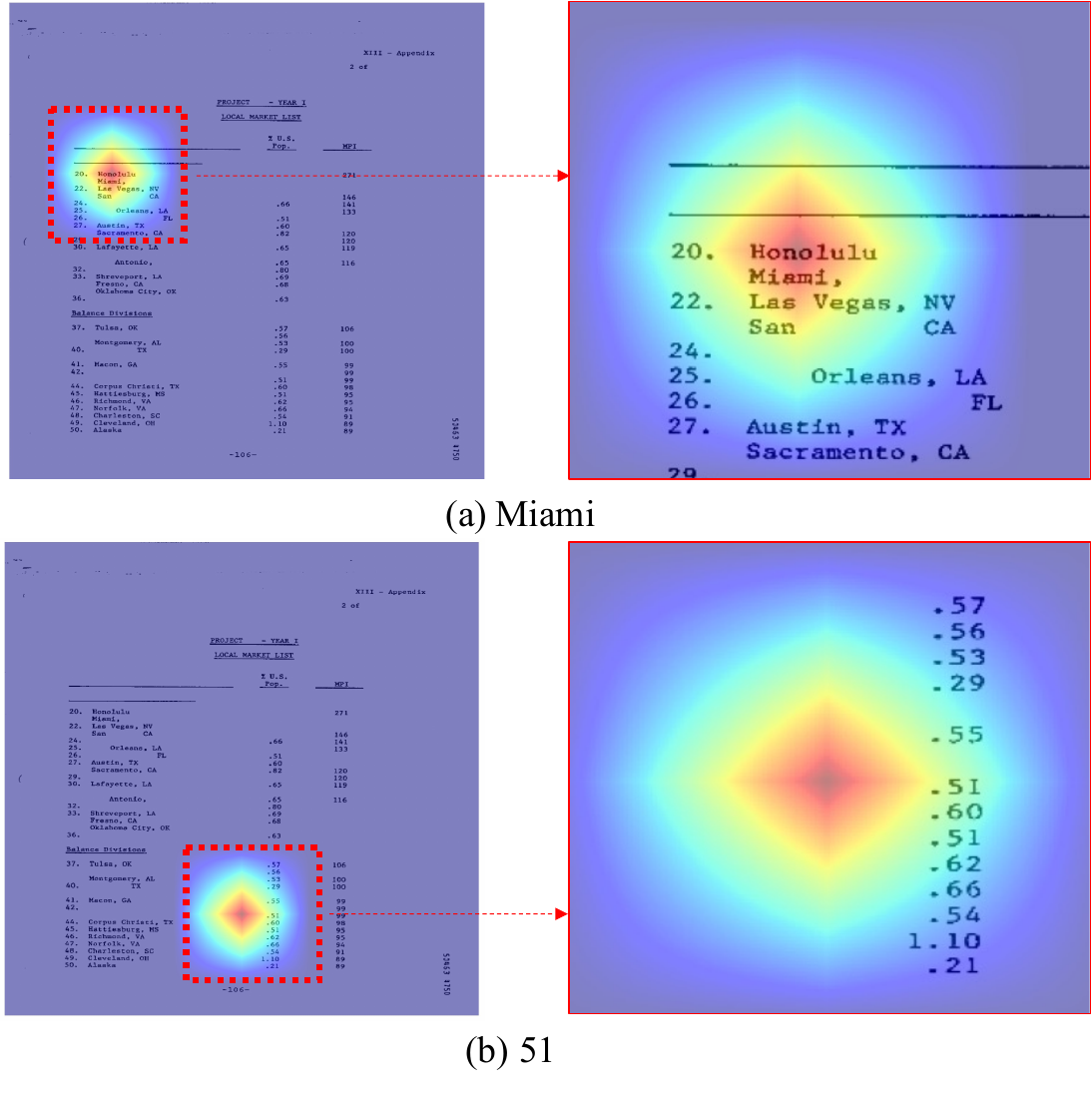}  
\caption{Visualization of the highest attention coefficient.}
\label{fig:t2i_attn}
\end{figure}

\subsection{ Comparisons with the SOTAs}
\textbf{Fine-tuning on Form Understanding} We treat the FUNSD as a sequence labeling task. Following previous works\cite{Elastic}, we set a token classifier on top of the text part of our model to predict BIO tags. As  mentioned in Section~\ref{sec:inputembedding}, the token local position in text blocks is a key information for the model to distinguish the beginning, inside, and outside of a text block.

As for fine-tuning settings, the learning rate of Bi-VLDoc is set to 1e-5. The learning rate of the classifier and the transition matrix is set to 1e-3. We fine-tune the pre-trained Bi-VLDoc on the FUNSD training set for 100 epochs with batch size 16.

Table~\ref{tab:funsd} shows the experimental results of entity-level precision, recall, and F1 scores on the FUNSD test set. As can be seen, the multi-modal models outperform the text-only models, and our Bi-VLDoc model outperforms  all previous multi-modal pre-trained models by a large margin; an improvement of 8.3\%  for the F1 score. This illustrates that our bidirectional vision-language information with layout pre-training strategy is quite beneficial to document understanding.

\textbf{Fine-tuning on Receipt IE} Following previous works, we use Bi-VLDoc to extract key information within a piece of text on the CORD dataset, which is treated as a sequence labeling task. Above the text part of the output features from the Bi-VLDoc, we build a random initialized token-level classification layer to predict the BIO tags. We fine-tune the pre-trained Bi-VLDoc on the CORD training set for 200 epochs. The batch size is set to 16. The learning rate is 1e-5. 

Table~\ref{tab:cord} presents the experimental results on the CORD test set. Bi-VLDoc significantly outperforms the multi-modal models and the text-modality models. The best CORD performance is improved from 0.9601 to 0.9784 which demonstrates the effectiveness of the Bi-VLDoc model.

\textbf{Fine-tuning on Document Classification} In document classification, Bi-VLDoc is required to categorize documents into one of the specified classes. We leverage the multimodal features in this downstream task. We build a classification layer above the [CLS] output feature from cross attention to predict the document classes. We fine-tune the pre-trained Bi-VLDoc on the RVL-CDIP training set for 200 epochs. The batch size is set to 16. The learning rate is 1e-5. 

Table~\ref{tab:rvlcdip} gives the classification accuracy on the RVL-CDIP test set. The experiment results have shown that all the text, image, and layout information are important to the document classification task. The best performance is achieved by the Bi-VLDoc, where an improvement of 1.04\% is observed compared to the current SOTA results. This illustrates that the bidirectional vision-language information with the layout in Bi-VLDoc leads to better understanding of documents, thus resulting in new SOTA on the document classification task.

\textbf{Fine-tuning on Document Visual QA} For document visual question answering task, given the question, the Bi-VLDoc is required to find a short span of answer from the given context. It is formalized as an extractive QA task. We set question tokens, context tokens and visual tokens to segment 0, 1 and 2, respectively. A token-level classifier on top of the text part is built to do span extraction. Inspired by previous works on DocVQA, we first use the train\&dev dataset to do continue pre-training. Then, following \cite{xu2020layoutlmv2} and \cite{li2021structurallm}, we build a Question Generation(QG) dataset that contains almost one million question-answer pairs generated by a QG model from IIT-CDIP. Finally, we fine-tune Bi-VLDoc with batch size 16 and learning rate 1e-5 on the DocVQA train\&dev set.

The DocVQA Average Normalized Levenshtein Similarity (ANLS) scores of previous text-only, layout-only, multi-modal, and the DocVQA top-1 models are listed in Table ~\ref{tab:docvqa}. Compared to the multi-modal pre-trained document model LayoutLM v2, with the same modal information, our Bi-VLDoc achieved an improvement of over 1.26\% ANLS. Bi-VLDoc shows better performance over existing single model methods on the leaderboard. This also verifies that the pre-trained Bi-VLDoc model not only benefits the traditional text IE task and image classification task, but also the high-level document understanding VQA task through the effective bidirectional vision-language information with layout model training.

\subsection{Visualizations}


In order to further understand the bidirectional vision-language interaction, we compare the LayoutLM v2 \cite{xu2020layoutlmv2} with our methods. In Figure \ref{fig:vis}, textual features (upper part) and visual features (lower part) are concatenated to a unified sequence. It can be seen that the attention scores in LayoutLM v2 are radial, which means that only visual features can be aware of the textual features. The attention scores in our proposed Bi-VLDoc are cross, which means that both visual features and textual features can be aware of each other.

Some visualization examples of the visual backbone from Bi-VLDoc are shown in Figure \ref{fig:t2i_attn}, we evaluate the attention coefficient of each token for the visual region, and show the region with the highest attention coefficient. It is noted that each different text token receives the highest attention from the visual area corresponding to its semantic content in the heat map. This demonstrates that the proposed model can learn visual representations of the texture information.

\begin{figure*}[htbp]
\centering  
\includegraphics[scale=0.65]{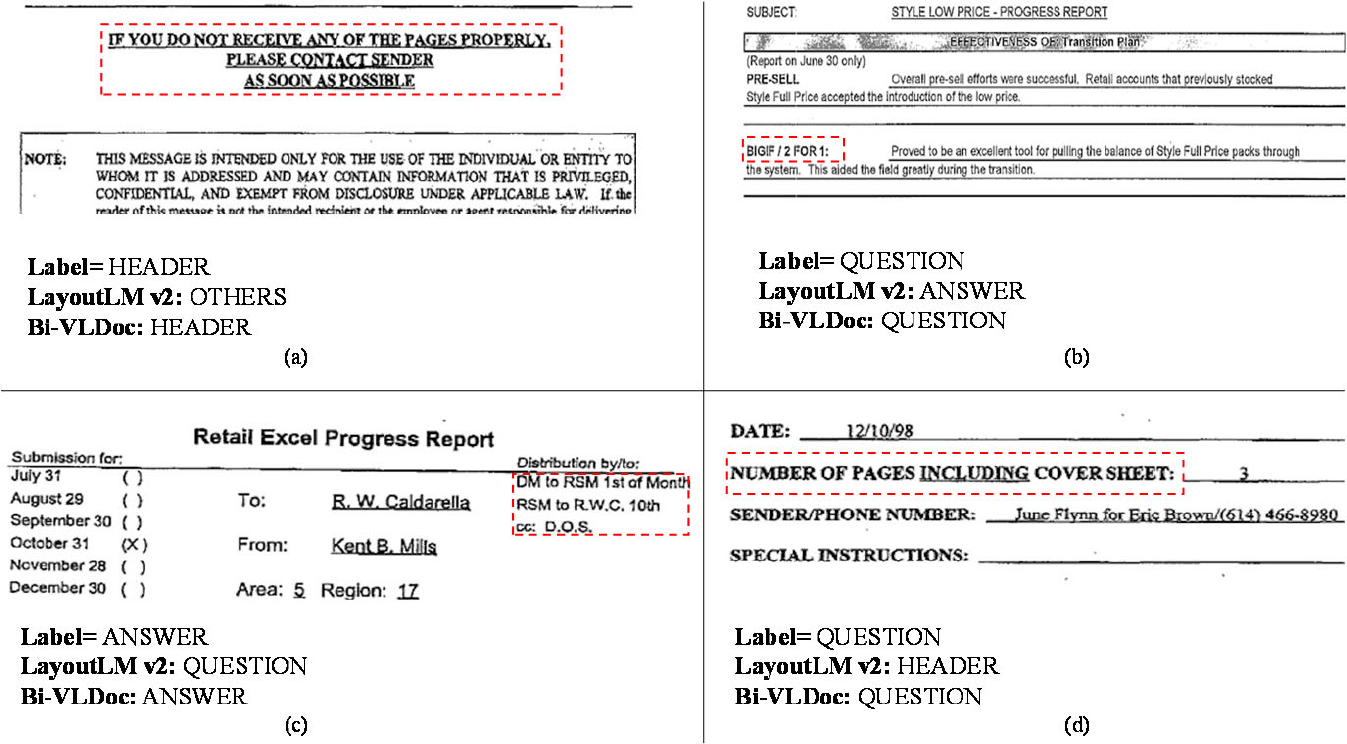}  
\caption{Examples of the output of LayoutLM v2 and Bi-VLDoc on the FUNSD dataset.}
\label{fig:case_study}
\end{figure*}

\subsection{Case Study}
\label{sec:casestudy}

Some examples of the output of LayoutLM v2 \cite{xu2020layoutlmv2} and Bi-VLDoc on the FUNSD test set are shown in Figure \ref{fig:case_study}. 
As stated above, compared with LayoutLM v2, Bi-VLDoc improves vision-language interactions. 
Take the image on the top-left of Figure \ref{fig:case_study} as an example. 
In this example, the model needs to label the tokens in the red frame with the HEADER entity. The result of LayoutLM v2 is OTHERS. 
The corresponding visual content of all the tokens in this entity is in a notable visual feature which is in the middle of the document and is in bold font.
Therefore, Bi-VLDoc predicts the correct result with the ability to map the text tokens to their corresponding visual content.

\section{Conclusion}

This paper proposes Bi-VLDoc, a task-agnostic representation learning framework for VrDU. A novel bidirectional vision-language hybrid-attention operation is introduced for multi-modal information alignment and fusion. By combining three new pre-training objectives (BTIA, RWTP, and TIPA), Bi-VLDoc achieve effective  bidirectional vision-language supervision. Experimental results on form understanding, receipt IE, document classification, and document VQA datasets confirmed that Bi-VLDoc significantly outperforms the previous SOTA methods in these downstream tasks. 
For future work, we plan to explore modeling long documents in cross-modal VrDU tasks.

\section{Limitation and Discussion}
Although the Bi-VLDoc method proposed in this paper explores a balanced and interactive modeling between semantics and vision, it still has the following shortcomings. Firstly, the Bi-VLDoc does not explore the downstream structure suitable for visual tasks, such as detection, recognition, layout analysis and so on. Secondly, the Bi-VLDoc proposed in this paper does not make a targeted design in the pre organization of reading 
order. The performance of our Bi-VLDoc still has room for further improvement.





\newpage

\end{document}